\documentclass[conference]{IEEEtran}
\IEEEoverridecommandlockouts
\usepackage{cite}
\usepackage{amsmath,amssymb,amsfonts}
\usepackage{algorithmic}
\usepackage{graphicx}
\usepackage{textcomp}
\usepackage{xcolor}
\usepackage{booktabs}
\usepackage{epsfig}
\usepackage{enumitem}
\usepackage[ruled,vlined]{algorithm2e}
\usepackage{multirow}
\usepackage{color,soul}
\usepackage{float}
\usepackage{hyperref}
\usepackage{graphicx}
\usepackage{amsmath}
\usepackage{amssymb}
\usepackage{booktabs}

\usepackage{amsfonts}
\usepackage{algorithmic}
\usepackage{textcomp}
\usepackage{xcolor}
\usepackage{epsfig}
\usepackage{enumitem}
\usepackage{multirow}
\usepackage{color,soul}
\usepackage{float}
\DeclareMathOperator*{\argmax}{arg\,max}
\DeclareMathOperator*{\argmin}{arg\,min}

\newcommand{\ignore}[1]{}
\makeatletter
\newcommand{\linebreakand}{%
  \end{@IEEEauthorhalign}
  \hfill\mbox{}\par
  \mbox{}\hfill\begin{@IEEEauthorhalign}
}
\makeatother

\def\BibTeX{{\rm B\kern-.05em{\sc i\kern-.025em b}\kern-.08em
    T\kern-.1667em\lower.7ex\hbox{E}\kern-.125emX}}
\begin{document}

\title{Cocktail Party Attack: Breaking Aggregation-Based Privacy in Federated Learning using Independent Component Analysis}

\author{\IEEEauthorblockN{Sanjay Kariyappa$^\dag$}
\IEEEauthorblockA{\textit{Georgia Institute of Technology} \\
sanjaykariyappa@gatech.edu}
\thanks{$\dag$work was done while the authors were at FAIR/Meta AI.}
\and
\IEEEauthorblockN{Chuan Guo}
\IEEEauthorblockA{\textit{Meta AI} \\
chuanguo@fb.com}
\and
\IEEEauthorblockN{Kiwan Maeng$^\dag$}
\IEEEauthorblockA{\textit{Pennsylvania State University} \\
kvm6242@psu.edu}
\and
\IEEEauthorblockN{Wenjie Xiong}
\IEEEauthorblockA{\textit{Meta AI / Virginia Tech} \\
wenjiex@fb.com}
\linebreakand
\IEEEauthorblockN{G. Edward Suh}
\IEEEauthorblockA{\textit{Meta AI / Cornell University} \\
edsuh@fb.com}
\and
\IEEEauthorblockN{Moinuddin K Qureshi}
\IEEEauthorblockA{\textit{Georgia Institute of Technology} \\
moin@gatech.edu}
\and
\IEEEauthorblockN{Hsien-Hsin S. Lee$^\dag$}
\IEEEauthorblockA{\textit{} \\
lee.sean@gmail.com}
}
\maketitle
\begin{abstract}

Federated learning (FL) aims to perform privacy-preserving machine learning on distributed data held by multiple data owners. To this end, FL requires the data owners to perform training locally and share the gradient updates (instead of the private inputs) with the central server, which are then securely aggregated over multiple data owners. Although aggregation by itself does not provably offer privacy protection, prior work showed that it may suffice if the batch size is sufficiently large. In this paper, we propose the Cocktail Party Attack (CPA) that, contrary to prior belief, is able to recover the private inputs from gradients aggregated over a very large batch size. CPA leverages the crucial insight that aggregate gradients from a fully connected layer is a linear combination of its inputs, which leads us to frame gradient inversion as a blind source separation (BSS) problem (informally called the cocktail party problem). We adapt independent component analysis (ICA)---a classic solution to the BSS problem---to recover private inputs for fully-connected and convolutional networks, and show that CPA significantly outperforms prior gradient inversion attacks, scales to ImageNet-sized inputs, and works on large batch sizes of up to 1024.
\end{abstract}

\section{Introduction}
Federated learning (FL)~\cite{mcmahan2017communication} is a powerful and flexible framework for privacy-preserving machine learning (ML) model training on distributed data. The FL framework typically consists of a central server and multiple clients that hold private training data. The protocol involves the server distributing the model parameters $\theta$ to the clients, and then the clients using this model to compute gradient update $\nabla_\theta \mathcal L$ using their private data. The gradient updates are aggregated and shared with the server who updates its model parameters using the aggregated gradients and this process is repeated until convergence. 

\begin{figure}[tb]
	\centering
    \centerline{\epsfig{file=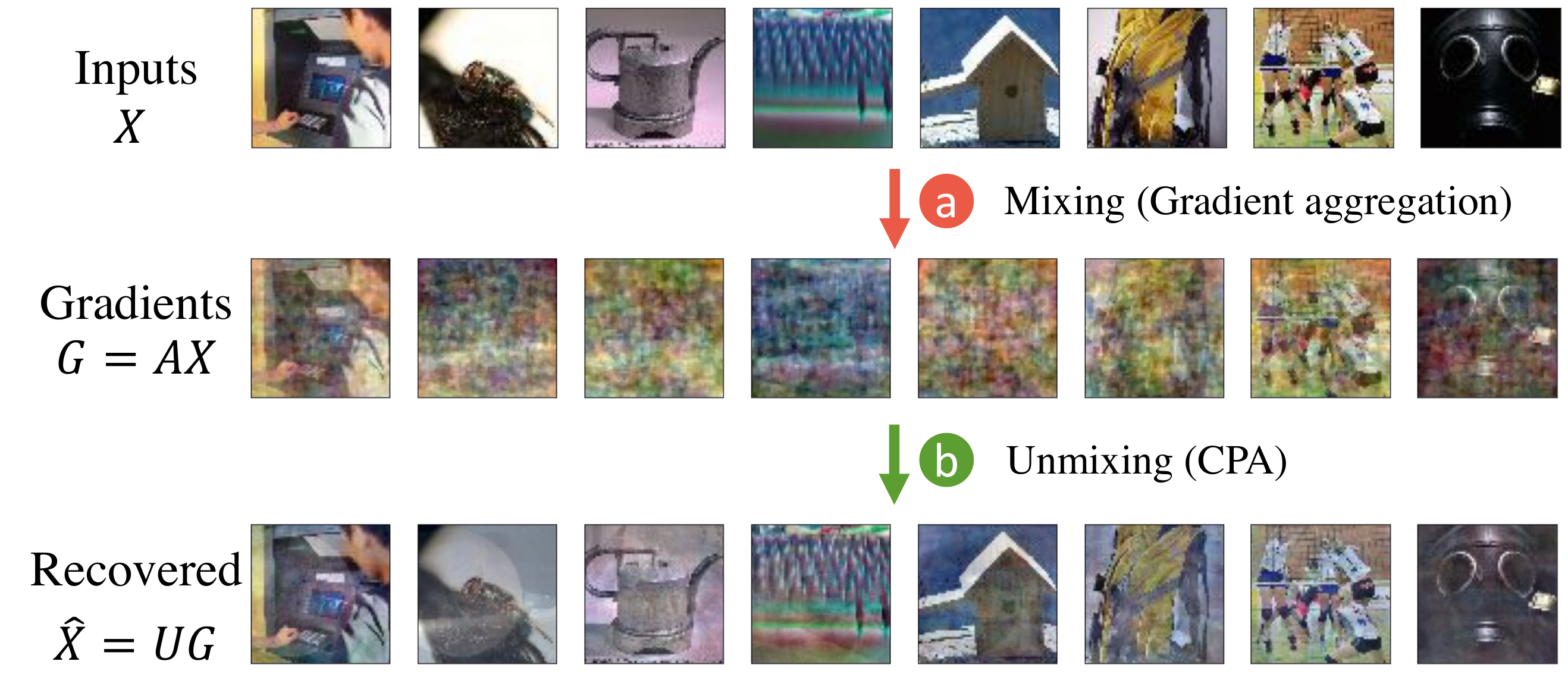, width=\columnwidth}}
	\caption{(a) Cocktail party attack (CPA) is based on the insight that aggregate gradients from FC layers are linear combinations of its inputs. (b) CPA uses this insight to frame gradient inversion as a blind source separation problem and recovers the inputs from the gradients by optimizing an unmixing matrix $U$ using independent component analysis.}
	\label{fig:summary}
\end{figure}

\ignore{
\begin{figure}[tb]
	\centering
    \centerline{\epsfig{file=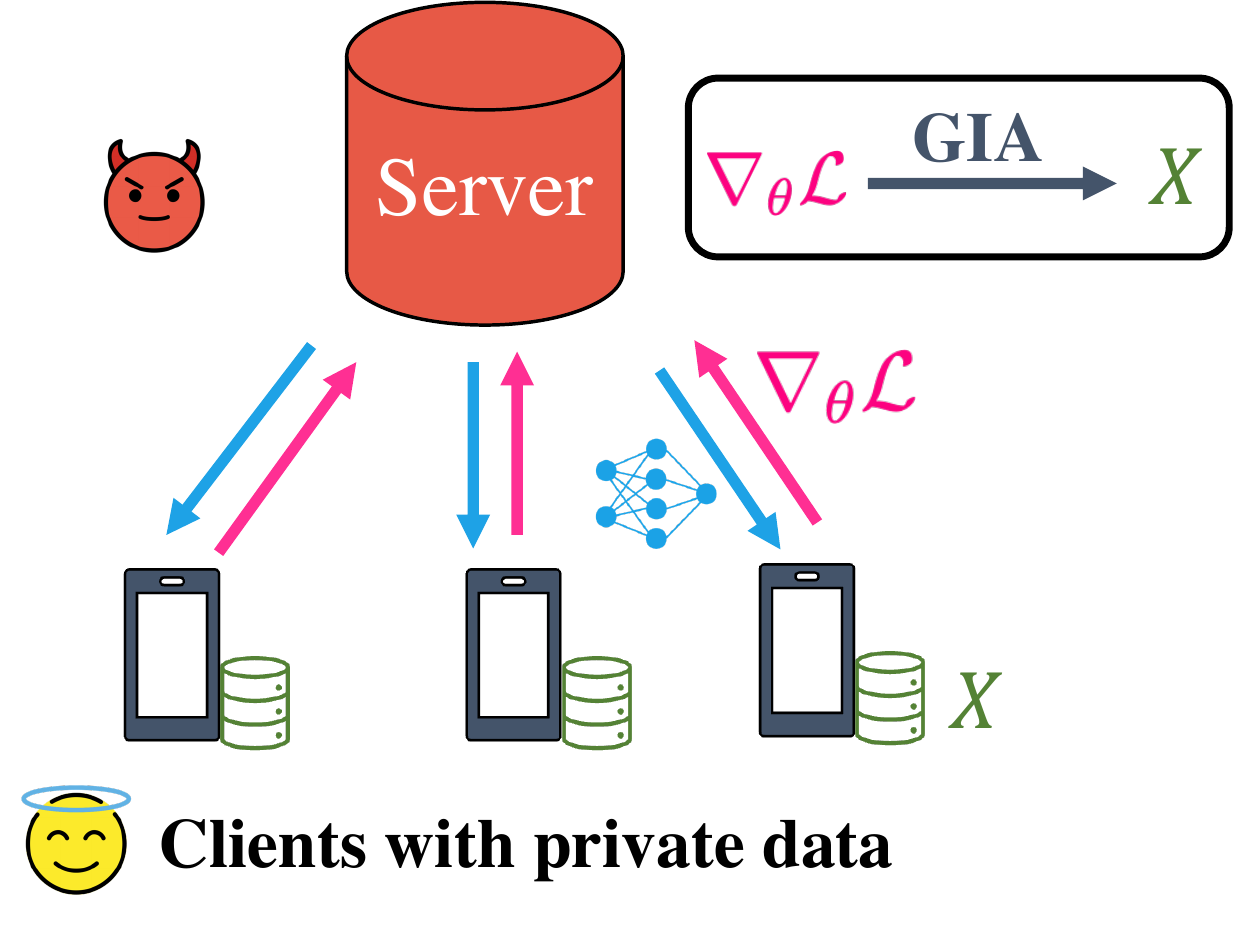, width=0.65\columnwidth}}
	\caption{In FL, the clients perform training locally on their private data $X$ and send the weight updates/gradients $\nabla_\theta \mathcal L$ to the server. Gradient inversion attacks (GIA) try to break privacy by recovering the private data from the gradients.}
	\label{fig:fl}
\end{figure}
}

While FL avoids the direct sharing of data, this in itself does not guarantee privacy as the gradient update shared with the server can contain information about the private training data~\cite{nasr2019comprehensive}. For instance, Zhu et al.~\cite{dlg} showed concretely that these gradient updates can be \emph{inverted} to recover their associated private data in a process now called \emph{gradient inversion}. In realistic settings, however, gradient inversion in FL is considered to be hard~\cite{dlg}, as gradients are aggregated across a large number of inputs (e.g., 1024), which thwarts most existing gradient inversion attacks.
\ignore{
\begin{figure*}[tb]
	\centering
    \centerline{\epsfig{file=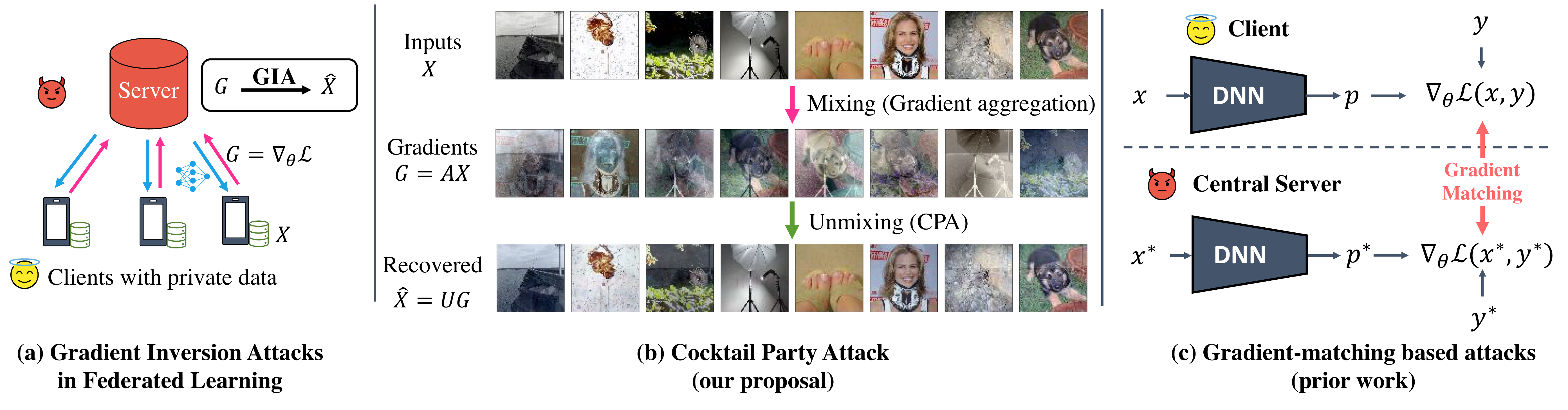, width=\textwidth}}
	\caption{(a) Gradient inversion attacks (GIA) break privacy by recovering the private data from the aggregate gradients in federated learning. (b) Cocktail party attack (CPA) leverages the observation that aggregate gradients from fully connected layers are linear combinations of its inputs. It optimizes an unmixing matrix $U$ using independent component analysis to recover the private inputs from the aggregate gradients. (c) Prior works optimize a set of dummy parameters ($x^*, y^*$) with the objective of matching the gradient obtained during FL to perform gradient inversion.}
	\label{fig:motivation}
\end{figure*}}

The central goal of our paper is to show that this hardness is not an inherent privacy property of FL, but rather a technical limitation of the existing attack algorithms. To this end, we propose the \emph{Cocktail Party Attack (CPA)}---a gradient inversion attack that is able to faithfully recover the inputs to a fully-connected (FC) layer from its aggregated gradient. This is made possible by a novel insight that the aggregated gradient for an FC layer is a linear combination of its inputs, which allows us to frame the recovery of these inputs as a \emph{blind source separation} (BSS) problem and solve it using independent component analysis (ICA).

When applied to a fully-connected network (or any network where the first layer is an FC layer), CPA can readily perform gradient inversion to recover input images. Fig.~\ref{fig:summary} shows an illustration of CPA on a fully-connected network, where a batch of training images is faithfully recovered from its aggregated gradient. We further extend CPA to perform gradient inversion on convolutional networks by first recovering the per-sample embeddings to an FC layer of the network and then inverting these embeddings using feature inversion to recover the input images. Empirically, we show that CPA has the following advantages over prior work:
\begin{itemize}[leftmargin=*]
    \item We evaluate CPA by inverting the gradients from a FC network trained on CIFAR-10~\cite{cifar10} and Tiny-Imagenet~\cite{le2015tiny}, and a VGG-16 network~\cite{simonyan2014very} trained on ImageNet to show that our attack can perform high-quality recovery of private inputs even with batch size as large as 1024.
    \item Compared to prior work based on gradient matching~\cite{dlg}, CPA can recover inputs with better quality and scales to datasets with larger input sizes (e.g., ImageNet). Furthermore, we show that gradient matching can be combined with CPA to further improve attack performance.
    \item CPA only uses simple image priors such as smoothness and does not require knowledge of the input data distribution or changes to the model parameters, and hence is more versatile and applicable to real world settings.
\end{itemize}
The efficacy of CPA shows that aggregation alone does not provide meaningful privacy guarantees and defenses like differential privacy are truly necessary to prevent gradients from leaking private data in FL.
\ignore{
1. Prior works on graident inversion use gradient matching combined with additional regularization terms. We first describe the gradient matching technique and the variants of this technique
}

\section{Background}
In this section, we provide background on FL and gradient inversion attacks (GIA). We also provide an overview of prior works on GIA and describe their limitations.

\subsection{Federated Learning}
\begin{figure}[htpb]
	\centering
    \centerline{\epsfig{file=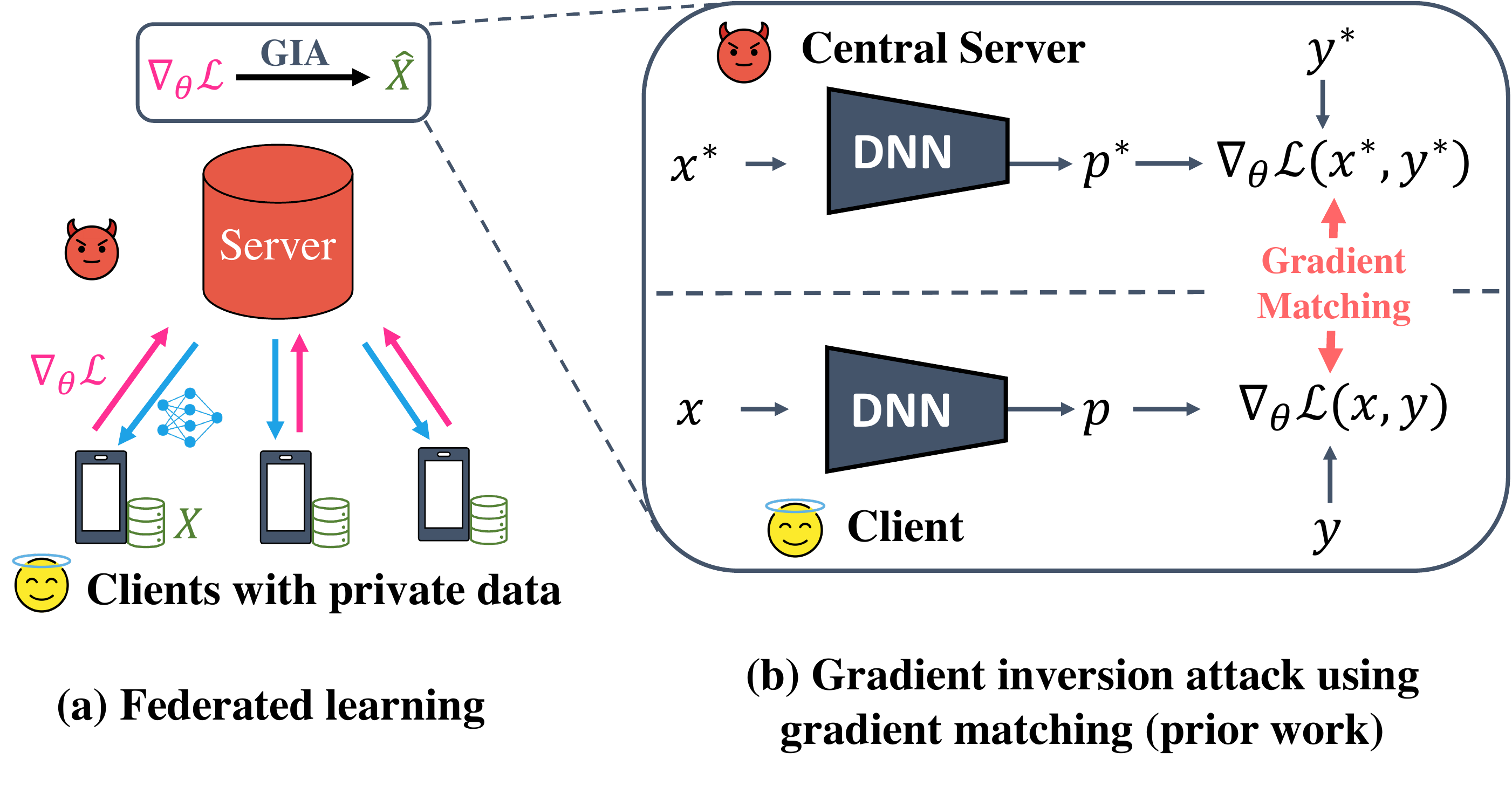, width=\columnwidth}}
	\caption[]{(a) FL aims to perform privacy preserving ML on distributed data by requiring the clients to perform training locally on their private data $X$ and send the weight updates/gradients $\nabla_\theta \mathcal L$ to a central server. (b) Gradient inversion attacks (GIA) break privacy by recovering the private data from the gradients. Prior works carry out GIA by optimizing a set of dummy parameters ($x^*, y^*$) with the objective of matching the gradient obtained during FL.\footnotemark}
	\label{fig:fl_gm}
\end{figure}

FL aims to train a model on distributed data held by multiple clients in a privacy-preserving manner. FL involves a central server and multiple clients who hold private data $X$ as shown in Fig.~\ref{fig:fl_gm}a. To train a model $f_\theta$, the server starts by distributing the model to the clients. The client uses the batch of private data to compute the aggregate gradients of the loss with respect to the model parameters $\theta$ and sends the gradient $\nabla_\theta\mathcal L$ to the central server. 
The server collects and aggregates the gradients, and uses the aggregated gradient to update the model parameters.
To improve privacy, some proposals use secure aggregation~\cite{secure_aggregation}. This process is repeated until the model converges. 

\subsection{Gradient Inversion Attack}

The goal of a gradient inversion attack is to recover the inputs from the aggregate gradients produced during training. Such attacks can be used by a malicious server to leak the private inputs of the clients in FL.

\textbf{Attack Objective:} Let $\mathcal{A}$ denote the gradient inversion attack. The goal of the attack is to estimate the batch of inputs $\hat{X}$ from the aggregate gradient $\nabla_\theta \mathcal L(S, Y)$, such that the estimated inputs $\hat{X}$ are semantically similar to the original inputs $X$ as shown in Eqn.~\ref{eq:attack}.  
\begin{align} \label{eq:attack}
\min d(\hat{X}, X), \text{where }\hat{X} = \mathcal{A}(\nabla_\theta \mathcal L(X, Y)) 
\end{align}
\textbf{Attack Constraints:} We assume an honest-but-curious adversary, meaning that the central server is not allowed to modify the FL protocol or insert malicious weights~\cite{curious, rob} to achieve the attack objective.

\footnotetext{All emojis in this paper are from https://openmoji.org/ and licensed CC BY-SA 4.0.}
\subsection{Related Work}
Most prior works on gradient inversion primarily rely on the \emph{gradient matching} objective to carry out the attack. We start by describing gradient matching, followed by attacks that build on this objective using various input priors like total variation prior, batch-norm statistics, and generative image priors to improve attack performance. We also discuss a recent line of work that uses malicious model parameters to carry out GIA. We describe the limitations of these prior works and also discuss the key challenge in scaling gradient matching based attacks to datasets with large inputs. 

\textbf{Gradient Matching:} Gradient matching~\cite{dlg} performs gradient inversion by optimizing a batch of dummy inputs and labels $(x^*, y^*)$ to produce a gradient that matches the one received by the server during FL as shown in Fig.~\ref{fig:fl_gm}b. This can be done by minimizing the distance between the gradient produced by the dummy variables $\mathcal L_\theta(x^*, y^*)$ and the gradient received during FL $\mathcal L_\theta(x, y)$ as shown in Eqn.~\ref{eq:gm}. This method was shown to work well on small datasets like CIFAR-10~\cite{cifar10} up to a batch size of 8.
\begin{align} \label{eq:gm}
\hat{x}, \hat{y} = \argmin_{x^*, y^*}d(\mathcal L_\theta(x^*, y^*), \mathcal L_\theta(x, y))
\end{align}
Here, $d$ denotes a distance metric between vectors like cosine similarity or $L_2$ norm. A subsequent work~\cite{idlg} proposed a method to infer the ground truth labels by examining the gradients of the last layer. Knowing the labels can help improve the quality of gradient inversion. However, this method only works when no two inputs in the batch belong to the same output class, limiting its applicability.

\textbf{Total Variation (TV) Prior:} Geiping et al.~\cite{geiping} proposed to use the TV prior~\cite{tv} as a regularization term along with the gradient matching objective. The TV prior (Eqn.~\ref{eq:tv}) penalizes high-frequency components in the input and encourages the optimization to find natural-looking images. With the TV prior, gradient matching can be scaled to work on a batch size of up to 100 for CIFAR-10 images and up to a small number of inputs for ImageNet.  
\begin{align}\label{eq:tv}
    \mathcal{R}_{TV}(x^*) = \mathbb E\big[|x^*_{i+1,j} - x^*_{ij}|\big] + \mathbb E\big[|x^*_{i,j+1} - x^*_{ij}|\big] 
\end{align}
\ignore{
\begin{align} \label{eq:tv}
\hat{s}, \hat{y} = \argmin_{s^*, y^*}1 - \frac{\langle\mathcal L_\theta(s^*, y^*), \mathcal L_\theta(s, y)\rangle}{||\mathcal L_\theta(s^*, y^*)|| ||\mathcal L_\theta(s, y)||} + \lambda \mathcal{R}_{TV}(s^*)
\end{align}
The first term is an alternative formulation of the gradient matching loss that uses cosine similarity to measure the distance between the gradients. The second term is the TV prior, which penalizes high-frequency components in the input $s^*$ and encourages the optimization to find natural-looking images that satisfy the gradient matching objective. 
}
\textbf{Batch Norm Statistics:} Several works~\cite{yin2021see, gradvit} have proposed to use the mean and variance of the activations captured by the batch norm (BN) layers as a prior to improve gradient inversion using the regularization term in Eqn.~\ref{eq:bn}. Here, $\mu_l(x^*)$ and $\sigma^2_l(x^*)$ represent the mean and variance of the activation produced by the input $x^*$ at the $l$-th BN layer. This regularization term encourages the optimization to find inputs that produce activations whose distribution matches the BN statistics. Evaluations from this work show that the BN prior can enable gradient inversion on ImageNet with a batch size of up to 48 examples.
\begin{align} 
\begin{split}\label{eq:bn}
\mathcal R_{BN}(x^*) = \mathbb{E}_l \Big[||\mu_l(x^*) - BN_l(mean)||_2 \\+ ||\sigma^2_l(x^*) - BN_l(var)||_2\Big]
\end{split}
\end{align}
A key limitation of this work is that it can only be used for models that use BN layers.
In a real-world FL, the model might not contain BN layers because BN layers often degrades accuracy with a non-IID data~\cite{niid_quagmire}, which is common in FL~\cite{fl_challenges}.
Furthermore, BN statistics can be used to perform model inversion to leak the training data directly from the model parameters~\cite{dreaming} (without the need for gradients), posing a concern about the premise of using networks with BN layers to train on private data.


\textbf{Generative Image Prior:} Instead of performing optimization in the space of inputs, a recent work~\cite{gip1} proposes to move the optimization to the smaller latent space of a generative model (G) to find an input $x^*=G(z^*)$ that satisfies the gradient matching objective as shown in Eqn.~\ref{eq:gip}. The reduced space of optimization and the prior induced by the generative model helps the attack scale to imagenet-scale datasets with a small batch size.
\begin{align} \label{eq:gip}
\hat{z}, \hat{y} &= \argmin_{z^*\in\mathbb{R}^k, y^*}d(\mathcal L_\theta(G(z^*), y^*), \mathcal L_\theta(x, y))
\end{align}
Note that this method requires the adversary to have access to in-distribution data or a generative model that is trained on in-distribution data. This may not be realistic in several settings (e.g. medicine and finance), where in-distribution data/generative model may not be available. Additionally, such methods may not work well under dataset shift.

\textbf{Malicious Parameters:} Several recent works~\cite{curious, rob, wen2022fishing} have proposed methods to leak private inputs by using malicious model parameters under the assumption of a dishonest central server. Such methods use weights that cause the aggregate gradient or the difference between two aggregate gradients to be predominantly influenced by a single input. However, under our threat model of an honest central server, malicious modifications to model parameters are not allowed. Thus, we do not consider such attacks in our work. 

\subsection{Challenges for Scaling to Large Input Sizes}
The optimization complexity of gradient matching poses a fundamental limitation to the scalability of prior works to datasets with large inputs. Most prior works (with the exception of generative image prior) perform optimization directly in the input-space. The size of this optimization problem is $\mathcal{O}(n \times d_{in})$, where $n$ is the batch size and $d_{in}$ denotes the dimensionality of the input. The difficulty of performing this optimization scales linearly with the dimensionality of the input, preventing gradient-matching based from scaling to high-dimensional inputs. 

In contrast, our proposed attack (CPA) has an optimization complexity $\mathcal{O}(n \times n$), which is independent of the input-dimensionality. This unique feature keeps the size of the optimization problem low and enables CPA to scale to ImageNet-sized datasets even with large batch sizes.
\begin{figure*}[tb]
	\centering
    \centerline{\epsfig{file=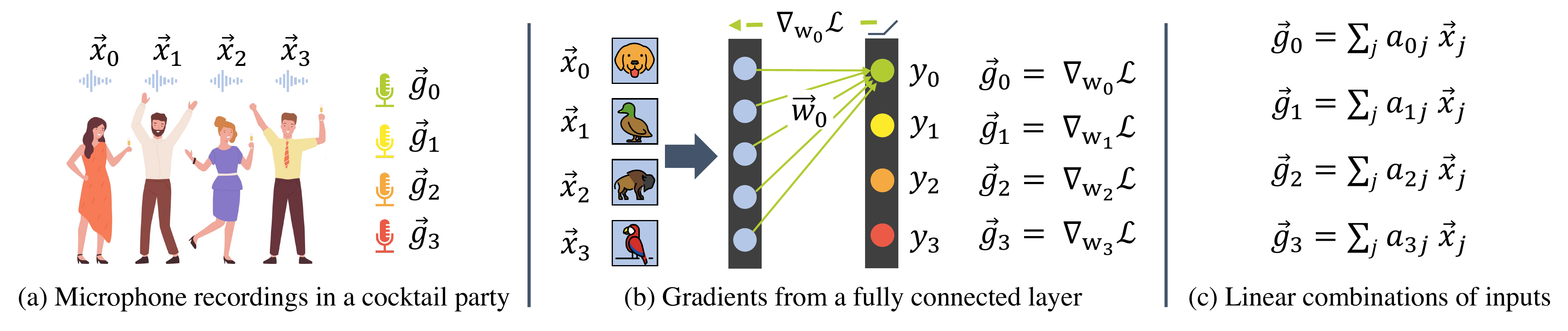, width=\textwidth}}
	\caption{The microphone recordings in the cocktail party problem  and the gradients from a fully connected layer can both be represented as linear combination of inputs. Recovering the inputs in both cases can be viewed as a blind source separation problem.}
	\label{fig:cocktail_party}
\end{figure*}
\section{Cocktail Party Attack}
Our paper proposes CPA -- a gradient-inversion attack specifically for FC layers. CPA frames gradient inversion as a BSS problem and uses ICA to recover the private inputs from aggregate gradients. We start by providing background on the BSS problem and describe how gradient inversion in FC layers can be viewed as a BSS problem. Next, we describe ICA--a classic signal processing technique-- that can be used to solve the the BSS problem. Lastly, we consider the problem of performing gradient inversion for FC models trained on image data, and describe how CPA can be used directly with simple image prior to perform gradient inversion and leak the private inputs. While FC models are typically not used for image classification, it represents the simplest setting to describe our attack. In the next section, we extend our attack to work on more realistic CNN models.

\subsection{Framing Gradient Inversion as a Blind Source Separation Problem}

CPA frames gradient inversion as a BSS problem. We explain the BSS problem using the motivating example of the cocktail party problem and show that gradient inversion for an FC layer can also be viewed as a BSS problem. 

\textbf{Cocktail Party Problem:} Consider a cocktail party where there are a group of four people talking simultaneously as depicted in Fig.~\ref{fig:cocktail_party}a. A microphone placed near this group picks up an audio recording consisting of an overlapping set of voices from the four speakers. We can model this recording by viewing it as a linear combination of the voices of the four speakers. More formally, if $\vec{x}_0, \vec{x}_1, \vec{x}_2, \vec{x}_3$ denote the speech vectors of the four speakers and $\vec{g}_0, \vec{g}_1, \vec{g}_2, \vec{g}_3$ denotes the recordings of four microphones places at different points, the recording of the $i$-th microphone is given by:
\begin{align} \label{eq:cp_equations}
\vec{g}_i = a_{i0} \vec{x}_0 + a_{i1} \vec{x}_1 + a_{i2} \vec{x}_2 + a_{i3} \vec{x}_3
\end{align}
Here, $a_{ij}$ denote the unknown mixing coefficients. The BSS problem can be stated as follows: given the mixed signals $\{\vec{g}_i\}$, recover the individual source signals $\{\vec{x}_i\}$.

\textbf{Gradient Inversion for FC Layer:}
Much like the cocktail party problem, the aggregate gradients from an FC layer can be viewed as linear combinations of the inputs used to generate them. To demonstrate this, consider an FC layer with four hidden neurons $y_0, y_1, y_2, y_3$ as shown in Fig.~\ref{fig:cocktail_party}b. Let $\vec{w}_0, \vec{w}_1, \vec{w}_2, \vec{w}_3$ represent the weight vectors associated with each output neuron. Let $\vec{x}_0, \vec{x}_1, \vec{x}_2, \vec{x}_3$ be a batch of four inputs used to perform a single iteration of training. $\vec{g}_i = \nabla_{w_i}\mathcal L$ is the aggregate gradient of the loss with respect to $\vec{w}_i$, which is computed by taking the mean of the individual gradients $\nabla_{w_i}\mathcal L^j$ associated with each input $\vec{x}_j$. This aggregate gradient $\vec{g}_i$ can further be expressed as a linear combination of the inputs $\vec{x}_j$ as follows:
\begin{align} 
\vec{g}_i &= \frac{1}{4}\sum_j\nabla_{w_0}\mathcal L^j = \frac{1}{4}\sum_j \frac{\partial \mathcal L}{\partial y^j_i} \frac{\partial y^j_i}{\partial w_i} = \frac{1}{4}\sum_j \frac{\partial \mathcal L}{\partial y^j_i}\vec{x}_j
\label{eq:gradient_linear_combination}
\end{align}
Notice that the aggregate gradients here have a 1-to-1 correspondence with the cocktail party problem. The inputs here are similar to the speakers and gradients are similar to the recordings of the microphones. Recovering the inputs $\{\vec{x}_i\}$ (source signals) from a set of aggregate gradients $\{\vec{g}_i\}$ (mixed signals) can thus be viewed as a BSS problem.

\textbf{Solving BSS with an Unmixing Matrix:} Eqn.~\ref{eq:gradient_linear_combination} can be expressed as a matrix multiplicaion operation as follows: $G = AX$. The rows of $X\in\mathbb R^{n\times d}$  denote the inputs (source signals), rows of the $A\in\mathbb R^{n\times n}$ represent the coefficients of linear combination and rows of the $G\in\mathbb R^{n\times d}$ denote the gradients (aggregate signals). We can estimate the source matrix $\hat{X}$ from $G$ by using an \emph{unmixing matrix} $U$ as follows: $\hat{X} = UG$. Each row of $\hat{X}$ represents a single recovered source signal $\hat{x}_i$. Note that $X$ can be recovered perfectly if $U=A^{-1}$ i.e. $\hat{X}\rightarrow X$ as $U\rightarrow A^{-1}$. Estimating $\hat{X}$ can thus be reduced to finding the unmixing matrix $U$.

\subsection{Gradient Inversion using ICA}
Independent component analysis (ICA) is a classic signal processing technique that can be use to solve the BSS problem by estimating the unmixing matrix $U$. To do this, ICA starts with a randomly initialized unmixing matrix $U^*$ and optimizes it to enforce certain properties on the recovered source signals. To explain, let $x^*_i=u^*_iG$ denote the $i$-th source signal recovered from multiplying the $i$-th row of $U^*$ with $G$. Note that the source signals represent images in our case (Fig.~\ref{fig:cocktail_party}b). ICA optimizes $U^*$ so that the recovered source signals $\{x^*_i\}$ satisfy the following key properties:
\begin{itemize}
    \item \emph{Non-Gaussianity:} Values of real-world signals such as images and speech typically do not follow a Gaussian distribution. We can measure non-Gaussianity using the negentropy metric, which can be estimated using Eqn.~\ref{eq:negentropy}~\cite{hyvarinen2000independent}. A high value of negentropy indicates a high degree of non-Gaussianity.
\begin{align}\label{eq:negentropy}
    J(x^*) = \mathbb E\Big[\frac{1}{a^2}\log\cosh^2(ax^*_i)\Big].
\end{align}
\item \emph{Mutual Independence (MI):} We assume that the source signals are independently chosen and thus their values are uncorrelated. Since each row $u^*_i$ of the unmixing matrix corresponds to a recovered source signal $x^*_i$, we can enforce MI by minimizing the absolute pairwise cosine similarity between the rows of $U^*$.
\begin{align}\label{eq:mi}
    \mathcal R_{MI} = \underset{i\neq j}{\mathbb E} exp\Big(T |CS(u^*_i, u^*_j)|\Big).
\end{align}
\item \emph{Source Prior:} Any prior information about the source signals can be included in our optimization in the form of an additional regularization term $\mathcal R_P$. 
\end{itemize}
We find the unmixing matrix $U$ by solving an optimization problem that combines the above properties.\footnote{We whiten and center the gradients before using it in our optimization.}$^,$\footnote{To understand the relative importance of different terms in the optimization, we refer the reader to the ablation study in Appendix~\ref{app:ablation}.}
\begin{align}\label{eq:ica}
\begin{split}
    U = \argmax_{U^*} \underset{i}{\mathbb E}\Big[J(u^*_iG) - \lambda_{P} \mathcal{R}_{P}(u^*_iG))\Big] - \lambda_{MI}\mathcal R_{MI} 
\end{split}
\end{align}

The $U$ matrix obtained from solving Eqn.~\ref{eq:ica} can be used to estimate the source matrix as follows: $\hat{X} = UG$. 

\subsection{CPA for FC Models} 
For an FC model trained on image data, where the inputs are directly fed to an FC layer,  we can recover the inputs directly by inverting the gradients of the first FC layer using CPA. Since the source signals are images, we can use TV regularization $\mathcal R_{TV}$ (Eqn.~\ref{eq:tv}) as the source prior in Eqn.~\ref{eq:ica}. One caveat of ICA is that it does not recover the sign of the input (i.e. $\hat{x}_i$ can be a sign-inverted version of $x_i$). However, this can be easily resolved by selecting between $\hat{x}_i$ and $-\hat{x}_i$ through a visual comparison.
\section{Extending Cocktail Party Attack to CNNs}
\label{sec:cpa_cnn}
\begin{figure}[tb]
	\centering
    \centerline{\epsfig{file=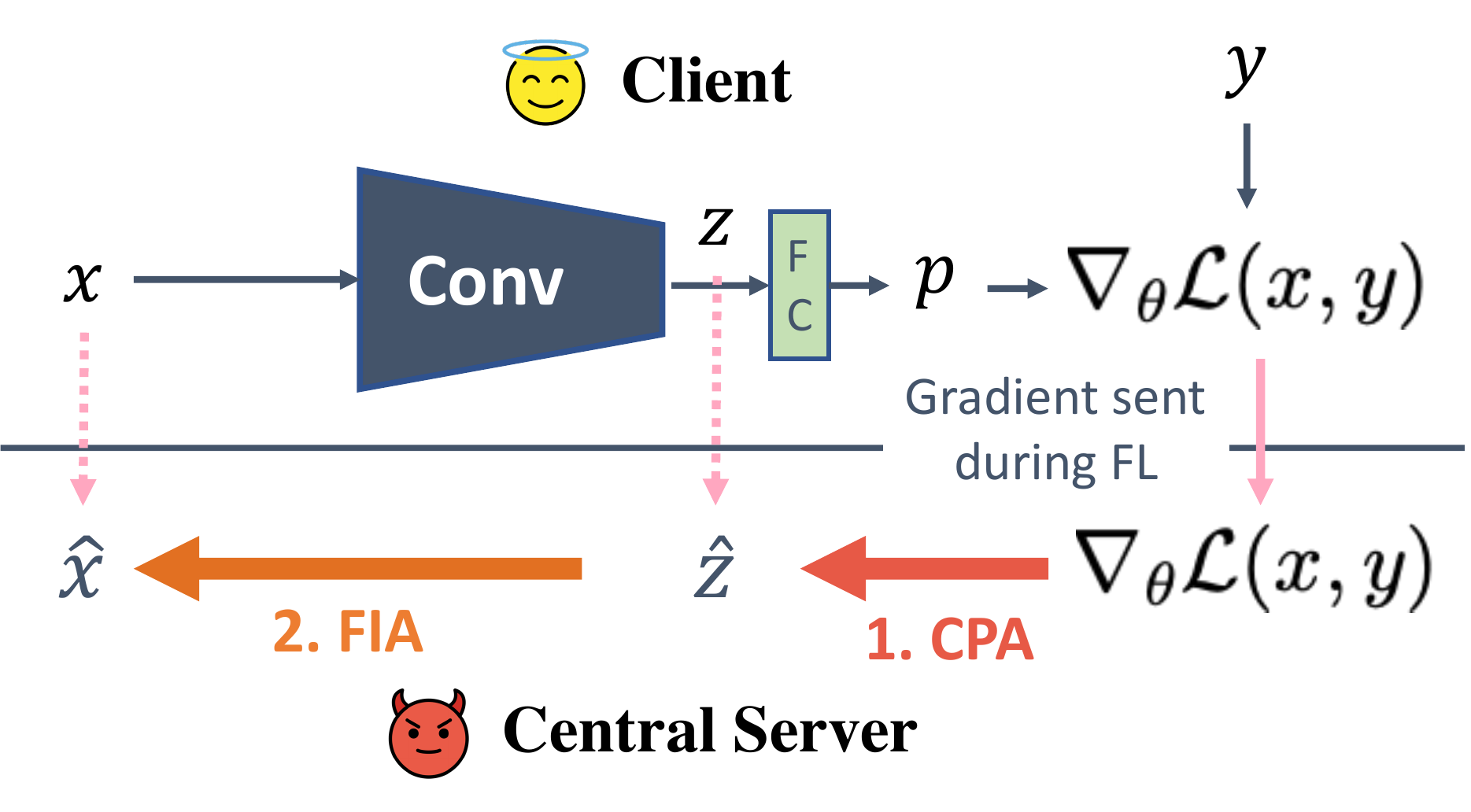, width=0.9\columnwidth}}
	\caption{We attack CNN models by first recovering the private embedding from the FC layer using CPA and then using a feature inversion attack (FIA) to recover the input from these embeddings.}
	\label{fig:cpa_fi}
\end{figure}

Since our attack is only applicable to FC layers, CPA cannot be used directly on CNN models, where the input is first passed through convolutional layers before being fed to a FC layer. However, we can use CPA in composition with a feature inversion attack (FIA) to recover the inputs (as shown in Fig.~\ref{fig:cpa_fi}) with the following two-step procedure:
\begin{enumerate}
    \item Use CPA on the FC layer to recover the embeddings from the gradients.
    \item Use FIA to estimate the inputs from the embeddings.
\end{enumerate}
We describe these two steps in greater detail below.
\ignore{
\begin{table*}[htb]
\centering
\begin{tabular}{ccccccccc}
\toprule
\multirow{2}{*}{\textbf{Attack}} & \multicolumn{8}{c}{\textbf{Batch Size}}                                                                             \\
\cmidrule{2-9}
                                 & 8              & 16             & 32             & 64             & 128            & 256            & 512   & 1024  \\
\midrule
GMA                              & 0.489          & 0.528          & 0.536          & 0.594          & 0.609          & 0.652          & -     & -     \\
CPA+FI                           & 0.501          & 0.51           & 0.483          & 0.493          & 0.479          & 0.495          & 0.507 & 0.509 \\
CPA+FIA+GMA                      & \textbf{0.413} & \textbf{0.466} & \textbf{0.392} & \textbf{0.430} & \textbf{0.423} & \textbf{0.469} & -     & -    \\
\bottomrule
\end{tabular}
\end{table*}
}
\subsection{Leaking Private Embeddings using CPA}
The gradients from the FC layer can be viewed a linear combinations of the embeddings $z$ that act as the input to the FC layer. We can use CPA to invert the gradients from the FC layer ($G$) and recover an estimate of the embeddings $\hat{z}=UG$. However, we can no longer use TV prior in our optimization (Eqn.~\ref{eq:cp_equations}) to find $U$ since the signal being recovered ($z$) is not an image. Instead, we use the following properties of embeddings produced with a ReLU non-linearity to design regularization terms for our optimization:
\begin{itemize}
\item \emph{$z$ is sparse:} Embeddings produced by networks that use ReLU non-linearity are sparse, as ReLU squashes negative activations to 0. We can use the $L_1$-norm: $|z^*|_1$ in our optimization to encourage sparsity of the estimated embeddings.
\item \emph{$z$ is a non-negative vector:} Values of the embedding vector are non-negative as ReLU truncates negative inputs to 0. We can minimize the following regularization function $\mathcal R_{NN}(z^*)=ReLU(-z^*)$ to encourage $z^*$ to be non-negative. However, the embedding recovered by ICA can be sign inverted. In this sign-inverted setting, the embedding can be non-positive vector. To allow for sign inverted recovery, we propose the \emph{sign regularization} (SR) function: $\mathcal R_{SR} = \min(ReLU(z^*), ReLU(-z^*))$. Minimizing $\mathcal R_{SR}$ ensures that $z^*$ is either non-negative or non-positive.
\end{itemize}
We combine the above regularization terms with the non-Gaussianity and mutual indepencence assumptions to derive the final optimization objective to estimate the unmixing matrix $U$ as follows:
\begin{align}\label{eq:ica_cnn}
\begin{split}
    U = \argmax_{U^*}\ &\underset{i}{\mathbb E} \Big[J(u^*_iG) - \lambda_{SP} |u^*_iG|_1 \\
    &- \lambda_{SR} \mathcal R_{SR}(u^*_iG)\Big] - \lambda_{MI}\mathcal R_{MI}.
\end{split}
\end{align}
\begin{figure*}[t]
	\centering
    \centerline{\epsfig{file=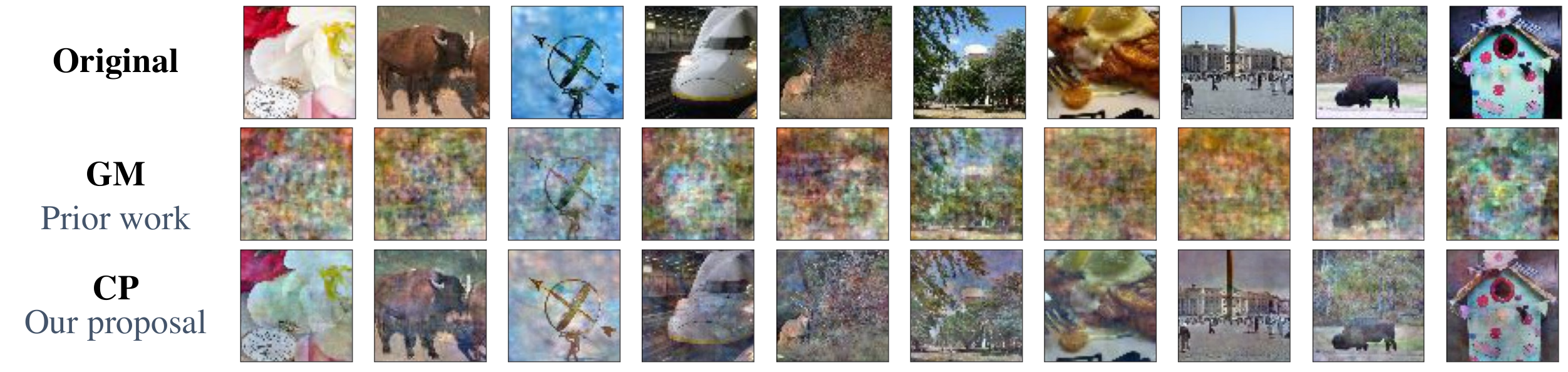, width=\textwidth}}
	\caption{Comparison of a subset of images recovered from gradient matching (GM) and cocktail party (CP) attacks by inverting the gradients from the FC-2 network with a batch of 64 Tiny-ImageNet inputs. The quality of images recovered by CP is significantly better than the GM attack. Please see Appendix~\ref{app:additional_results} for additional results.}
	\label{fig:results_tiny_imagenet}
\end{figure*}
The unmixing matrix can be used to recover the private embeddings $\hat{Z}$ from the gradient $G$ as follows: $\hat{Z}=UG$. We can use these private embeddings to recover the input using a feature inversion attack.
\subsection{Feature Inversion Attack}
FIA inverts the embedding produced by a neural network to recover the input. Formally, given a network $f:X\rightarrow Z$, and an embedding $x$, FIA recovers an estimate of the input $\hat{x}$ from $z$. We do this by solving the following optimization problem using a dummy input $x^*$:
\begin{align}\label{eq:fi}
\begin{split}
    \hat{x} = \argmax_{x^*} CS(f(x^*), z) - \lambda_{TV}\mathcal{R}_{TV}(x^*)  
\end{split}
\end{align}
The first term maximizes the cosine similarity between the dummy input $z^*=f(x^*)$ and true embedding $z$. The second term is TV regularization, which suppresses high-frequency components. Solving this optimization problem allows us to estimate of the private inputs $\{\hat{x_i}\}$ from the embedding $\{\hat{z_i}\}$ recovered by CPA, which completes the gradient inversion attack. Additionally, we can also use the gradient information to improve FIA by including the gradient matching objective in the optimization as follows:
\begin{align}\label{eq:fi+gm}
\begin{split}
    \hat{x} = \argmax_{x^*} CS(f(x^*), z) +& \lambda_{GM} CS(\nabla_\theta \mathcal L(x^*), \nabla_\theta \mathcal L(x))\\ 
    - &\lambda_{TV}\mathcal{R}_{TV}(x^*).
\end{split}
\end{align}

\section{Experiments}
To demonstrate the efficacy of our attack, we evaluate our proposed attack on FC and CNN models trained on image classification tasks. While FC networks are typically not used for image classification, they allow us to demonstrate the efficacy of our attack in its simplest form. Our evaluations on the CNN model (VGG-16) demonstrates the utility of our attack in a more realistic problem setting.

\subsection{Setup}
\textbf{Model and Datasets:} For our experiments on the FC model, we use a simple 2-layer network (FC-2), with the following network architecture: $[Linear(256)-ReLU()-Linear(k)]$ for a $k$-class classification problem. We train FC-2 on the CIFAR-10~\cite{cifar10} and Tiny-ImageNet~\cite{le2015tiny} datasets for 20 epochs using the Adam~\cite{adam} optimizer with a learning rate of $0.001$.  We perform our CNN experiments with a pre-trained VGG-16 network~\cite{simonyan2014very} and the ImageNet~\cite{deng2009imagenet} dataset.


\textbf{Evaluation Methodology:} We evaluate gradient inversion attacks with the following batch sizes: $[8, 16, 32, 64, 128, 256]$ for the FC-2 model and $[32, 64, 128, 256, 512, 1024]$ for VGG-16. We perform evaluations by first sampling a batch of inputs $\{x_i\}$ from an unseen test set to generate the aggregate gradient $\nabla_\theta\mathcal L$. We then use different gradient inversion attacks to recover an estimate of the inputs $\{\hat{x}_i\}$ from the aggregate gradients and compare their performance. Since our experiments are on image data, we use the LPIPS score~\cite{lpips} to quantify the perceptual similarity between the original and recovered images, and use this to evaluate the attacks. We repeat the attack on 5 batches of data and report the average LPIPS scores in our results. We set the number of optimization rounds for all attacks to $25000$.

\textbf{Hyperparameter Tuning:} For all the hyperparameters ($\lambda_{TV}, \lambda_{MI}, T, \lambda_{SP}, \lambda_{SR}$), we sweep their values in the range $[0.00001, 10]$ using a single batch of inputs and pick the set of values that yield the best LPIPS score to carry out our attack. Note that the inputs used in the hyperparameter sweep are separate from the ones used to report our results.
\begin{figure*}[t]
	\centering
    \centerline{\epsfig{file=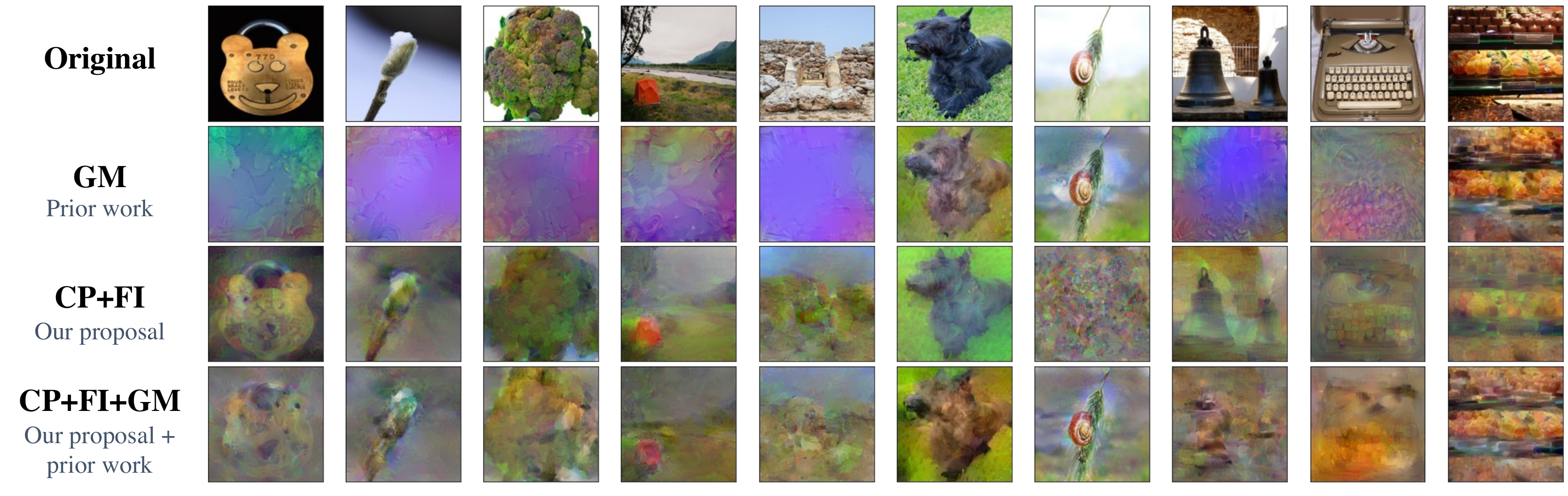, width=1.0\textwidth}}
	\caption{Comparison of a subset of images recovered from gradient matching (GM),  cocktail party + feature inversion (CP+FI) and cocktail party + feature inversion + gradient matching (CP+FI+GM) by inverting the gradients from a VGG-16 network with a batch of 256 ImageNet inputs. CP+FI (our proposal) can recover more images compared to GM (prior work). CP+FI+GM improves the quality of recovered images by combining the benefits of our proposal and prior work. Please see Appendix~\ref{app:additional_results} for additional results.}
	\label{fig:results_imagenet}
\end{figure*}
\textbf{Prior Work for Baseline:} Our threat model assumes an honest-but-curious attacker who does not have access to in-distribution examples\footnote{Except a single batch of inputs used to tune the hyperparameters}. The Geiping et al.~\cite{geiping} attack (which uses the gradient matching objective and TV prior) represents the stongest prior work under this setting\footnote{Attacks that use generative models~\cite{gip1} assume access to in-distribution data and cannot be used under our threat model}. We use this prior work (with the best choice of hyperparameters) as the baseline in our evaluations.

\begin{table}[tb]
\caption{$LPIPS\downarrow$ scores comparing the performance of cocktail party (CP) and gradient matching (GM) attacks on FC-2 trained on CIFAR-10 and Tiny-ImageNet. CP (our proposal) significantly outperforms GM (prior work) across all batch sizes.}
\centering
\scalebox{1.0}{
\begin{tabular}{ccccccc}
\toprule
\multicolumn{1}{c}{\multirow{2}{*}{\textbf{Attack}}} & \multicolumn{6}{c}{\textbf{Batch Size}}                                                             \\
\cmidrule(lr){2-7}
\multicolumn{1}{c}{}                                 & 8              & 16             & 32             & 64             & 128            & 256            \\
\midrule
\multicolumn{1}{c}{\textbf{}}                        & \multicolumn{6}{c}{\textbf{CIFAR-10}}                                                               \\
\cmidrule(lr){2-7}
GM                                                  & 0.283          & 0.390          & 0.491          & 0.569          & 0.610          & 0.614          \\
CP                                                  & \textbf{0.101} & \textbf{0.160} & \textbf{0.197} & \textbf{0.352} & \textbf{0.521} & \textbf{0.610} \\
\midrule
                                                     & \multicolumn{6}{c}{\textbf{Tiny-ImageNet}}                 
                                                     \\
\cmidrule(lr){2-7}
GM                                                  & 0.182          & 0.234          & 0.368          & 0.620          & 0.687          & 0.720         \\
CP                                                  & \textbf{0.082} & \textbf{0.143} & \textbf{0.164} & \textbf{0.217} & \textbf{0.232} & \textbf{0.388} \\
\bottomrule
\end{tabular}}
\vspace{-0.1in}
\label{table:results_cifar_tin}
\end{table}
\subsection{Results for FC-2}
We first present the results from our experiments on the FC-2 models trained on the CIFAR-10 and Tiny-ImageNet datasets. Fig.~\ref{fig:results_tiny_imagenet} shows qualitative results comparing CPA and GMA for Tiny-ImageNet with a batch size of 64. The images recovered by CPA have better quality and higher perceptual similarity with the original images, compared to the images recovered by GMA. Table~\ref{table:results_cifar_tin} shows quantitative results (LPIPS scores) comparing CPA and GMA with various batch sizes. A lower LPIPS value indicates better perceptual similarity and thus a better attack performance. Our result can be interpreted by considering the size of the optimization problem being solved by CPA and GMA.
\begin{itemize}
\item\emph{CPA} has an optimization complexity $\mathcal{O}(n\times n)$, as it is optimizing over $V^*$, which is an $n\times n$ matrix.
\item\emph{GMA} has an optimization complexity $\mathcal{O}(n\times d)$ as it is optimizing directly in the input space.  
\end{itemize}

Here, $n$ denotes batch size and $d$ denotes the dimensionality of the input ($d=3072$ for CIFAR-10 and $d=12288$ for Tiny-ImageNet). With this in mind, we make the following key observations from our results:
\begin{itemize}
\item \emph{Comparison with prior work:} CPA significantly outperforms GMA across all batches sizes since size of the optimization problem is much smaller for CPA compared to GMA. E.g. for n=64 with Tiny-ImageNet (d=12288), the size of the optimization is 4096 for CPA and 786432 for GMA.
\item \emph{Sensitivity to batch size:} The size of the optimization problem increases with an increase in batch size for both CPA and GMA causing their performance to degrade for larger batch sizes.
\item \emph{Sensitivity to input dimensionality:} The optimization in CPA is independent of the input dimensionality $d$. Consequently, CPA performs significantly better for datasets with larger inputs (Tiny-ImageNet) compared to GMA, especially for larger batch sizes.

\end{itemize}
\begin{figure}[tb]
	\centering
    \centerline{\epsfig{file=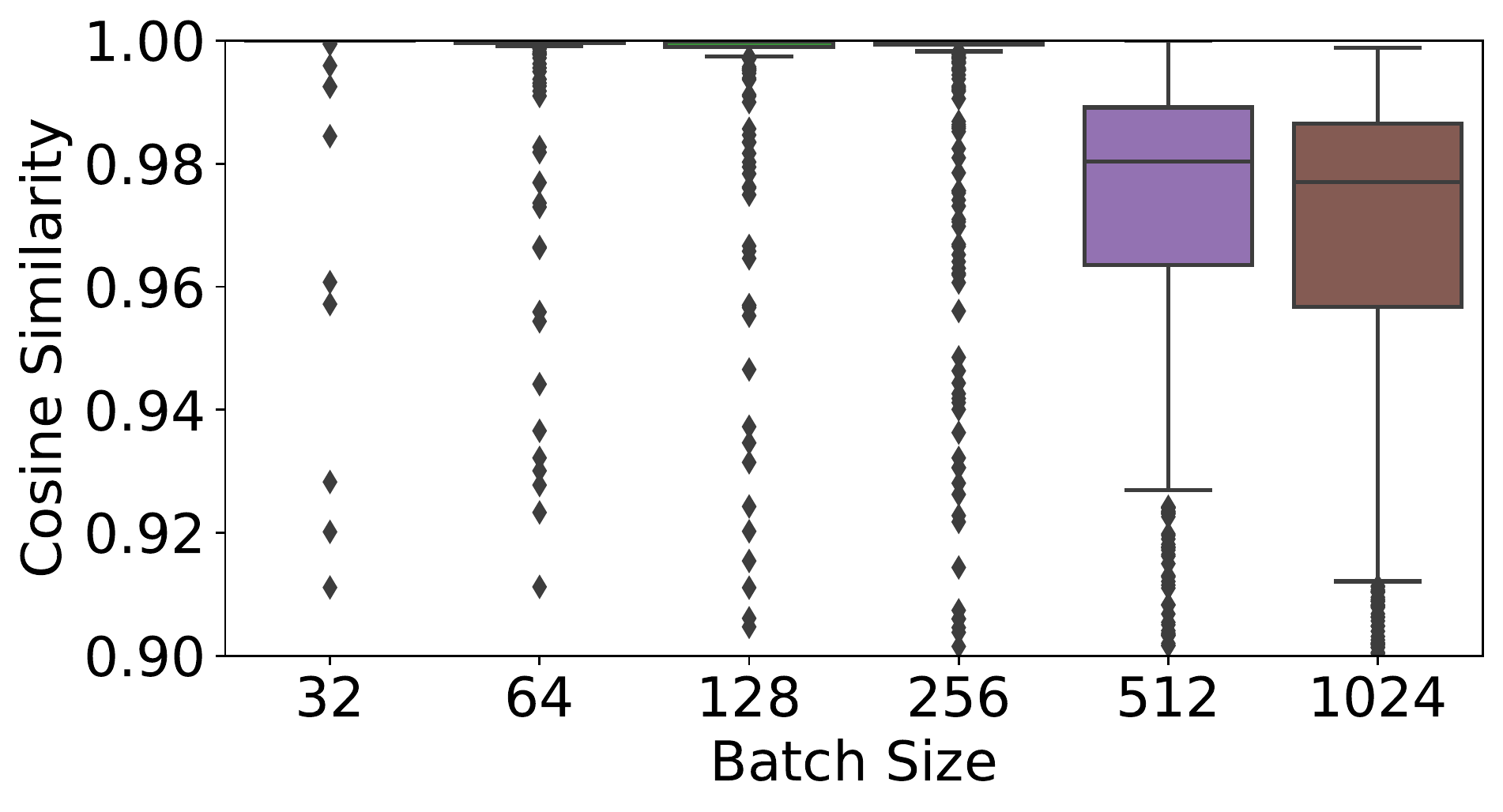, width=0.8\columnwidth}}
    \vspace{-0.1in}
	\caption{Distribution of cosine similarity (CS) values computed between the private embeddings $z$ and the embeddings recovered by CPA $\hat{z}$. Most of the CS values are close to the ideal value of 1.}
	\vspace{-0.1in}
	\label{fig:cs}
\end{figure}

\subsection{Results for VGG-16}
Next, we present the results from our experiments with the VGG-16 network trained on the ImageNet dataset. Our proposed attack uses a 2-step process that combines CPA and FIA (Fig.~\ref{fig:cpa_fi}) to perform gradient inversion. 

\textbf{Embedding recovery:} Our proposed attack starts by recovering the private embeddings $\hat{z}$ from the gradients of the FC layer. We evaluate the quality of these recovered embeddings by computing its cosine similarity (CS) with the original embedding $z$. Fig.~\ref{fig:cs} shows the distribution of the CS values for various batch sizes. Our results show that CPA allows near-perfect recovery of embeddings in most cases, with the CS values degrading slightly for larger batch sizes. 

\textbf{Gradient Inversion:} We use the embeddings recovered from CPA to estimate the inputs with a feature inversion attack. Table~\ref{table:results_imagenet} shows the $LPIPS$ scores comparing gradient matching attack (prior work), cocktail party + feature inversion attack (our proposal) and cocktail party + feature inversion + gradient matching attack (our proposal + prior work). We make the following key observations:

\begin{table}[b]
\caption{$LPIPS\downarrow$ scores of images recovered using GM (prior work), CP+FI (our proposal) and CP+FI+GM (prior work + our proposal) attacks, with VGG-16 network trained on ImageNet.}
\centering
\scalebox{1.0}{
\begin{tabular}{ccccccc}
\toprule
\multirow{2}{*}{\textbf{Attack}} & \multicolumn{6}{c}{\textbf{Batch Size}}                                           \\
\cmidrule{2-7}
                                 & 32             & 64             & 128            & 256            & 512   & 1024  \\
\midrule
GM                              & 0.536          & 0.594          & 0.609          & 0.652          & OOM     & OOM     \\
CP+FI                           & 0.483          & 0.493          & 0.479          & 0.495          & 0.507 & 0.509 \\
CP+FI+GM                      & \textbf{0.392} & \textbf{0.430} & \textbf{0.423} & \textbf{0.469} & OOM     & OOM\\
\bottomrule
\end{tabular}}
\vspace{-0.1in}
\label{table:results_imagenet}
\end{table}
\begin{itemize}
    \item \emph{Comparison with prior work:} CP+FI has better average $LPIPS$ score as it can recover more images compared to GM. CP+FI+GM improves the number of images recovered further by combining the befits of our proposal (CP+FI) and prior work (GM).
    \item \emph{Sensitivity to batch size:} The performance of GM degrades significantly with larger batch sizes. In contrast, CP+FI shows a smaller degradation and shows better scalability to larger batch sizes.
    \item \emph{Memory Footprint:} The optimization for gradient matching is $\mathcal{O}(n\times d)$. The memory footprint of this optimization can exceed the available GPU memory when the input dimensionality ($d$) is large. For ImageNet, we found that an 8-GPU machine cannot handle batch sizes in excess of 256 causing out of memory (OOM) errors. In contrast, the optimization for CPA is independent of $d$ and can scale to a batch size of 1024.
\end{itemize}

\section{Limitation and Defenses}
\textbf{Batch Size:} ICA requires the number of aggregate gradients from neurons (mixed signals) to be greater than or equal to the number of inputs (source signals). Thus choosing a very large batch size that exceeds the number of neurons in the FC layer can prevent our attack.

\textbf{Embedding size:} The efficacy of feature inversion attack depends on the size of the embedding. For a CNN that produces a smaller sized embedding, FIA might be harder to carry out. However, this limitation can be overcome if the attacker knows the input data distribution.

\textbf{Differential Privacy (DP) Defense:} Table~\ref{table:dp} shows the $LPIPS$ scores from CP and GM evaluated under DP noise~\cite{dwork2014algorithmic, dpsgd}. We use the FC2 model with Tiny-ImageNet dataset and a batch size of 8 for these experiments. We scale the gradients to have unit norm and perturb the gradients with different amounts of Gaussian noise. We also show the $\epsilon$ values for $(\epsilon, \delta)$ DP with $\delta=0.00001$ corresponding to different amounts of noise. Our evaluations show that DP noise provides an effective defense against our attack.
\begin{table}[htb]
\centering
\caption{$LPIPS\downarrow$ scores of recovered images from CP and GM attacks under varying magnitudes of DP noise.}
\scalebox{0.9}{
\begin{tabular}{ccccc}
\toprule
\textbf{$\sigma$}  & 0 & 0.0001 & 0.001 & 0.01 \\
\textbf{$\epsilon$} & $\infty$ & 6056.00         & 606.60         & 60.56         \\
\midrule
GM   & 0.182  & 0.426           & 0.728          & 0.701        \\
CP   & 0.0082  & 0.474           & 0.721          & 0.723         \\
\bottomrule
\end{tabular}
}
\label{table:dp}
\end{table}
\section{Extensions to CPA}
Our evaluations in this paper assume that the the network does not use batchnorm layers and that the attacker does not have access to the input data distribution. When this information is available, it can be combined with our attack to further improve performance. Additionally, the private embeddings leaked from CPA can also be used to infer additional attributes about the input~\cite{yeom2018privacy}. Lastly, our work can also be extended to language models and recommendation systems where it is common for the input to be fed directly to a FC layer. We leave this as part of our future work.

\section{Conclusion}
We propose \emph{Cocktail Party Attack (CPA)} -- a gradient inversion attack that can recovers private inputs from aggregate gradients in FL. Our work is based on the key insight that gradients from an FC layer are linear combinations of its inputs. CPA uses this insight to frame gradient inversion for an FC layer as a blind source separation problem and uses independent component analysis to recover the inputs. CPA can be used directly on FC models to recover the inputs. It can also by extended to CNN models by first recovering the embeddings from an FC layer and then using a feature inversion attack to recover the inputs from the embeddings. Our evaluations on several image classification tasks show that CPA can perform high-quality gradient inversion, scales to ImageNet-sized inputs, and works with a batch size as large as 1024. CPA is orthogonal to prior works that use gradient matching and can be combined with gradient matching based approaches to further improve gradient inversion. Our work demonstrates that that aggregation alone is not sufficient to ensure privacy and methods like differential privacy are truly necessary to provide meaningful privacy guarantees in FL.

{\small
\bibliographystyle{ieee_fullname}
\bibliography{references}
}
\clearpage
\appendix

\section{Ablation Study}
\label{app:ablation}
The optimization function used by CPA (Eqn.~\ref{eq:ica2}) consists of three terms that correspond to: 1. negentropy (NE) 2. total variation (TV) and 3. mutual independence (MI) objectives.
\begin{align}\label{eq:ica2}
\begin{split}
    \hat{V} = \argmax_{V^*} \underset{i}{\mathbb E} \Big[J(v^*_iG) - \lambda_{TV} \mathcal{R}_{TV}(v^*_iG)\\
    - \lambda_{MI}\underset{i\neq j}{\mathbb E} |exp(T\cdot |CS(v^*_i, v^*_j)|)|\Big]
\end{split}
\end{align}
To understand the importance of these three terms, we perform an ablation study. We use the FC2 model trained on TinyImagenet with a batch size 32 for our study and measure LPIPS by carrying out the attack by excluding different loss terms to understand their importance. We perform hyperparameter sweeps in each case and report the best (i.e. lowest) value of LPIPS in Table~\ref{table:ablation}. A higher value of LPIPS indicates a higher degradation in the quality of the image recovered, which implies a high level of importance on the term being removed. Our results indicate that the MI term is the most important. We find that without the MI term the optimization recovers the same image multiple times. TV is the second most important term, indicating that even a simple image prior is quite powerful. The NE term which enforces non-Guassianity has the lowest marginal benefit as it only provides a very weak prior on the source signal.
\begin{table}[htb]
\caption{Ablation study to understand the relative importance of different terms in the optimization function.}
\centering
\begin{tabular}{ccccc}
\toprule
      & NE+TV+MI & -NE & -TV & -MI    \\
\midrule
LPIPS $\downarrow$ &  0.081 &     0.092 & 0.368  &    0.546       \\
\bottomrule
\end{tabular}~\label{table:ablation}
\end{table}


\section{Additional Qualitative Results}
\label{app:additional_results}
\begin{figure*}[t]
	\centering
    \centerline{\epsfig{file=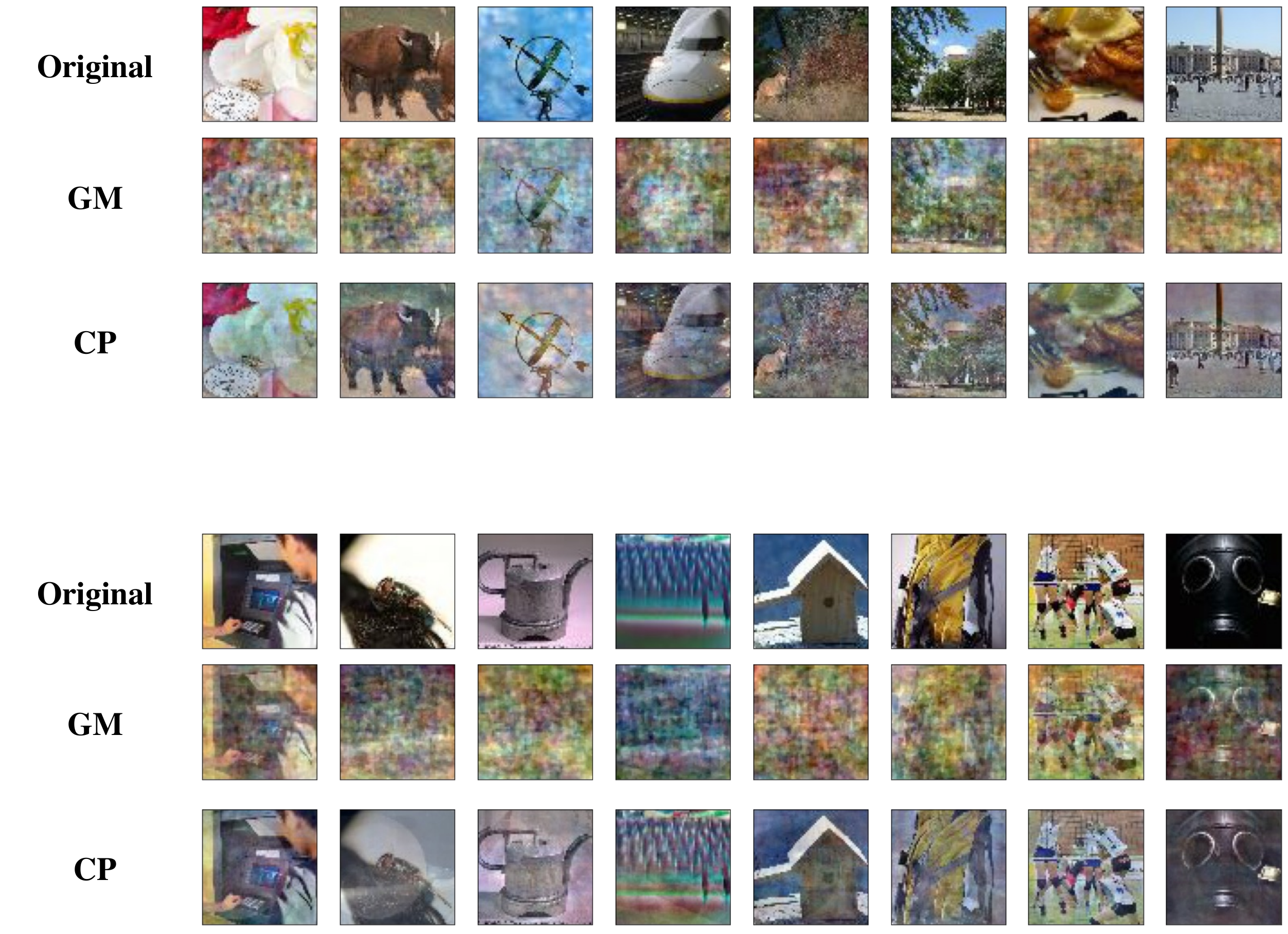, width=\textwidth}}
    \caption{Comparison of a random subset of images recovered by various gradient inversion attacks carried out using the gradients from FC2 with a batch of 64 Tiny-ImageNet images. Images are not cherry-picked.}
    \label{fig:tin1}
\end{figure*}

\begin{figure*}[t]
	\centering
    \centerline{\epsfig{file=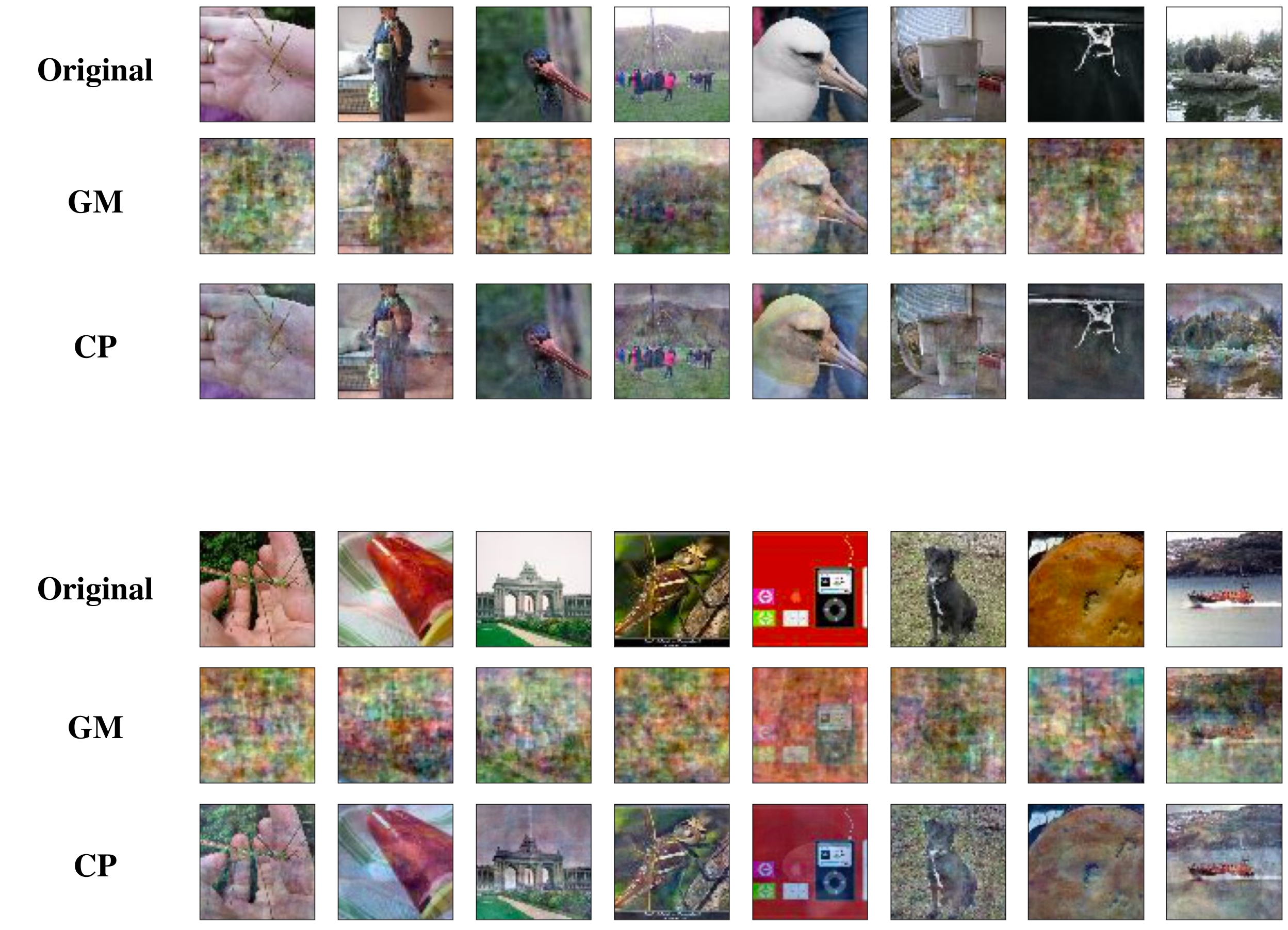, width=\textwidth}}
    \caption{Comparison of a random subset of images recovered by various gradient inversion attacks carried out using the gradients from FC2 with a batch of 64 Tiny-ImageNet images. Images are not cherry-picked.}
    \label{fig:tin2}
\end{figure*}

\begin{figure*}[t]
	\centering
    \centerline{\epsfig{file=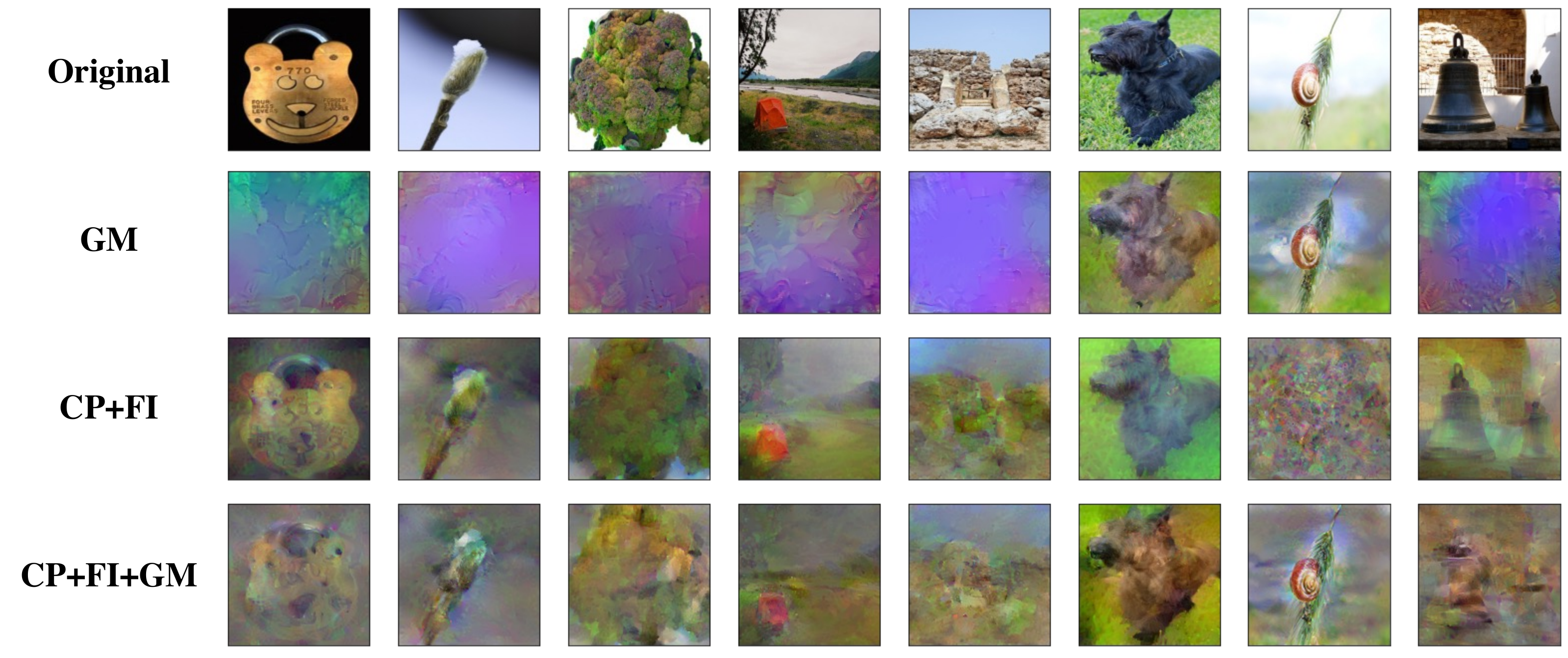, width=\textwidth}}
    \vspace{0.5in}
    \centerline{\epsfig{file=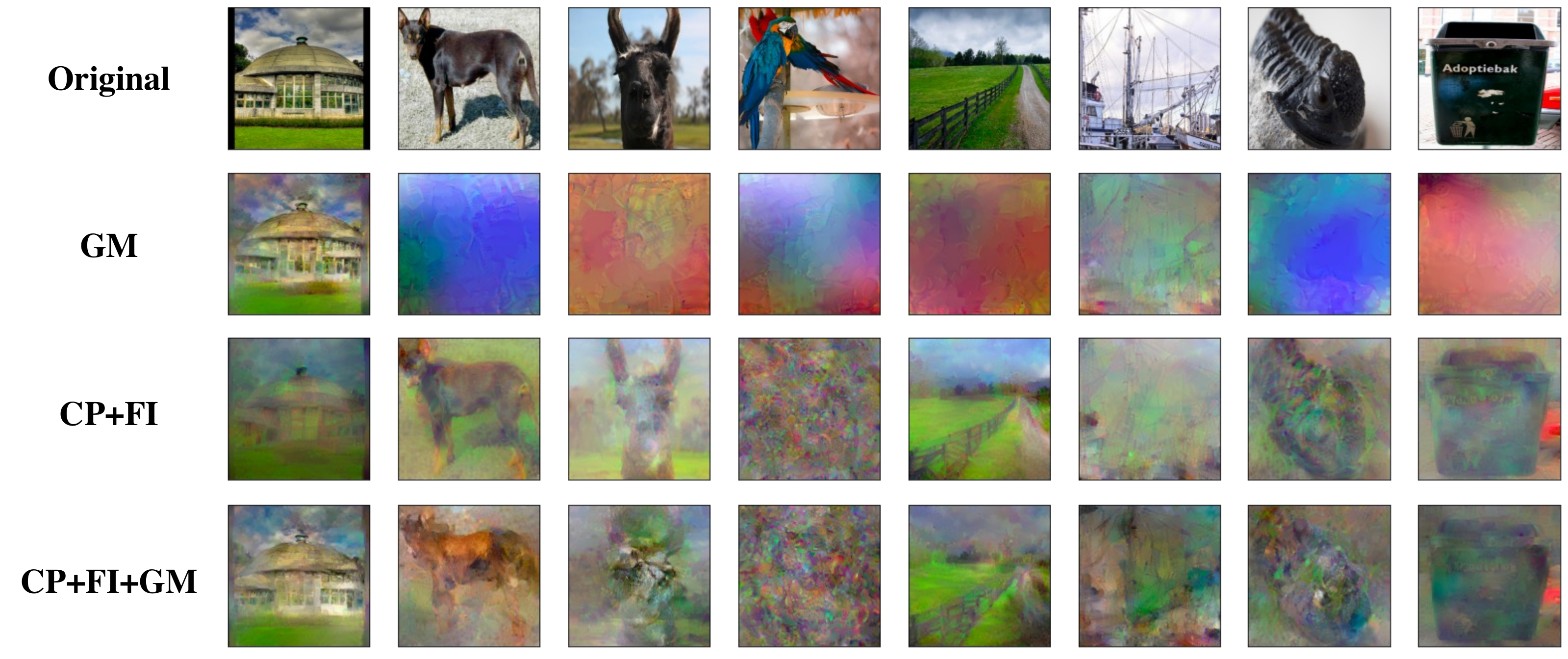, width=\textwidth}}
    \caption{Comparison of a random subset of images recovered by various gradient inversion attacks carried out using the gradients from VGG-16 with a batch of 256 ImageNet images. Images are not cherry-picked.}
    \label{fig:im1}
\end{figure*}

\begin{figure*}[t]
	\centering
    \centerline{\epsfig{file=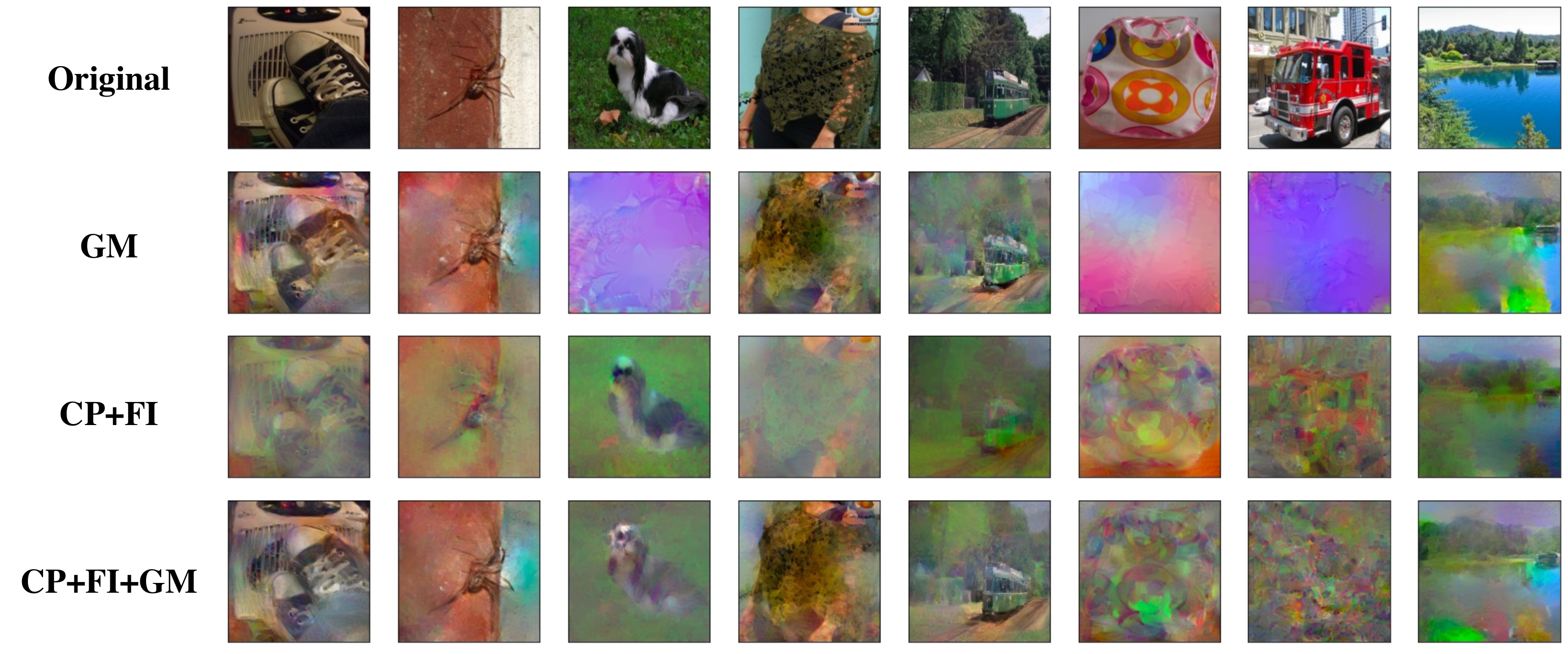, width=\textwidth}}
    \vspace{0.5in}
    \centerline{\epsfig{file=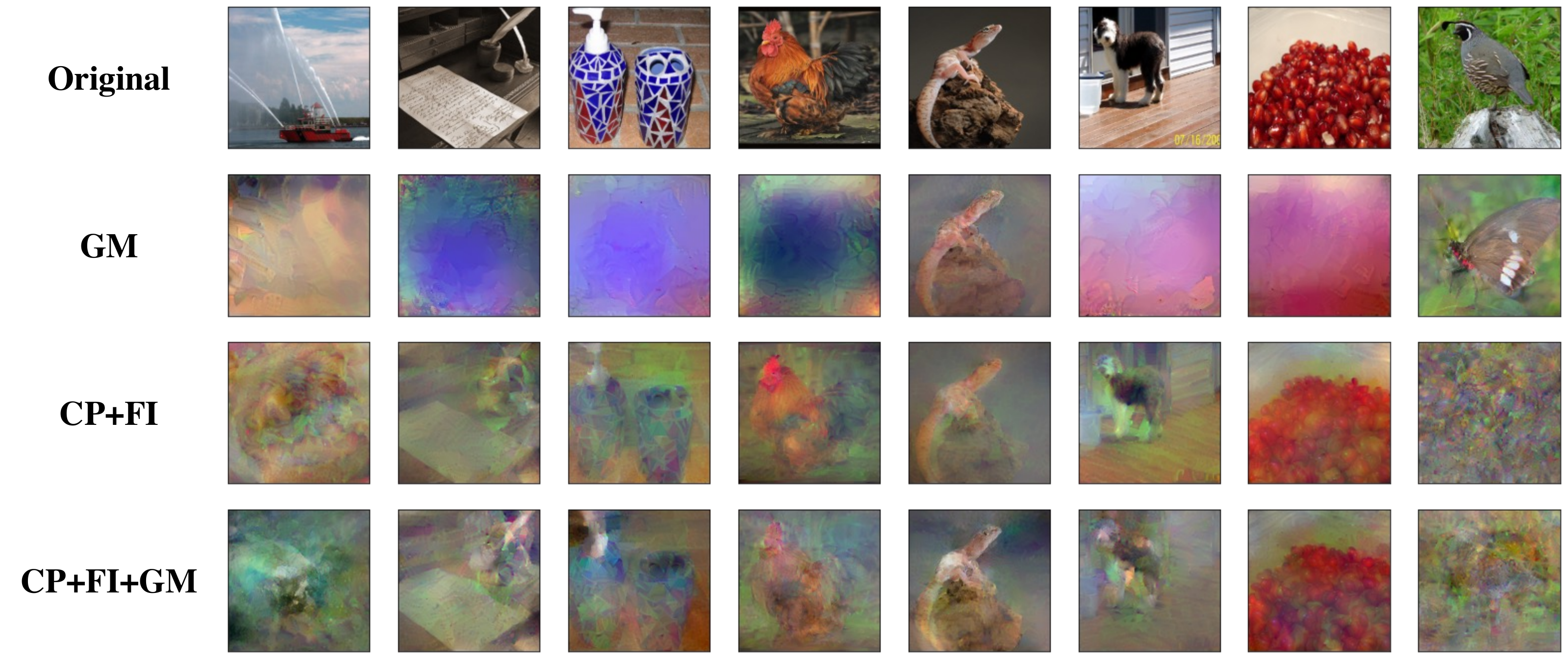, width=\textwidth}}
    \caption{Comparison of a random subset of images recovered by various gradient inversion attacks carried out using the gradients from VGG-16 with a batch of 256 ImageNet images. Images are not cherry-picked.}
    \label{fig:im2}
\end{figure*}

Fig.~\ref{fig:tin1}, Fig.~\ref{fig:tin2}, Fig.~\ref{fig:im1} and Fig.~\ref{fig:im2} show additional qualitative results comparing the recovered images from gradient inversion attacks on Tiny-ImageNet and ImageNet.

\ignore{\begin{figure*}[t]
	\centering
    \centerline{\epsfig{file=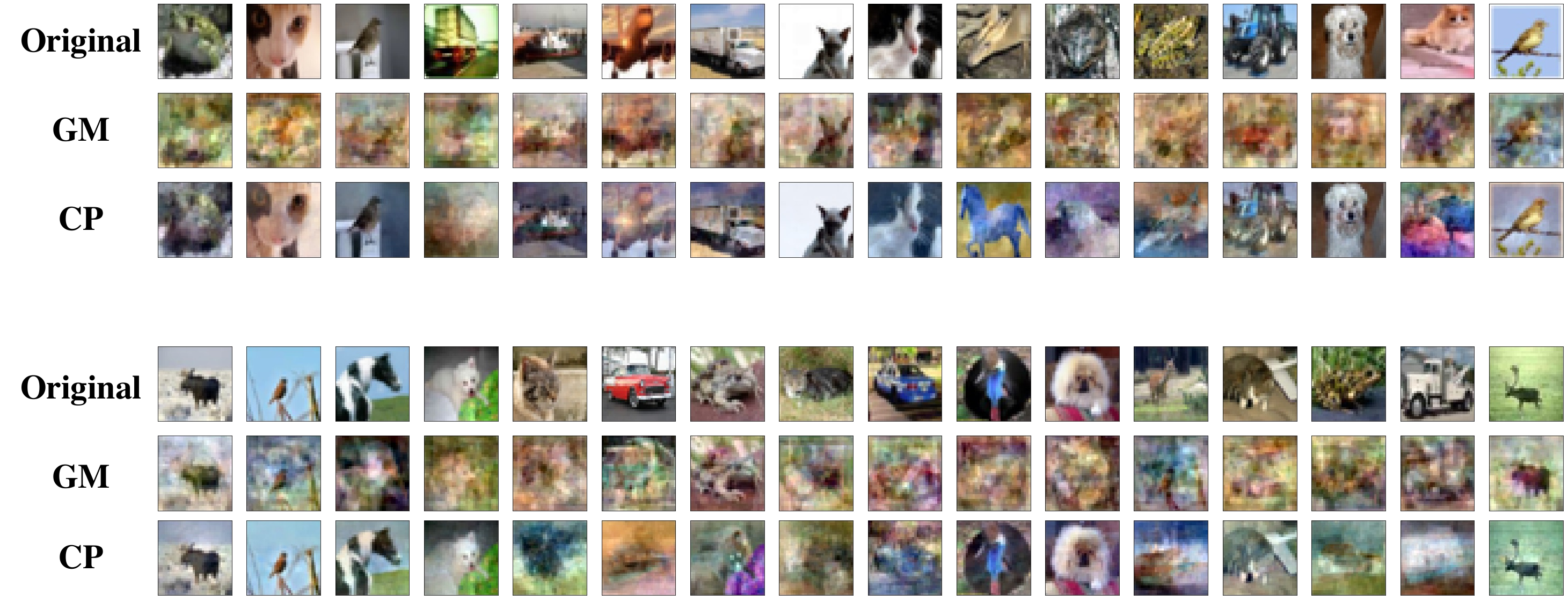, width=\textwidth}}
    \vspace{0.4in}
    \centerline{\epsfig{file=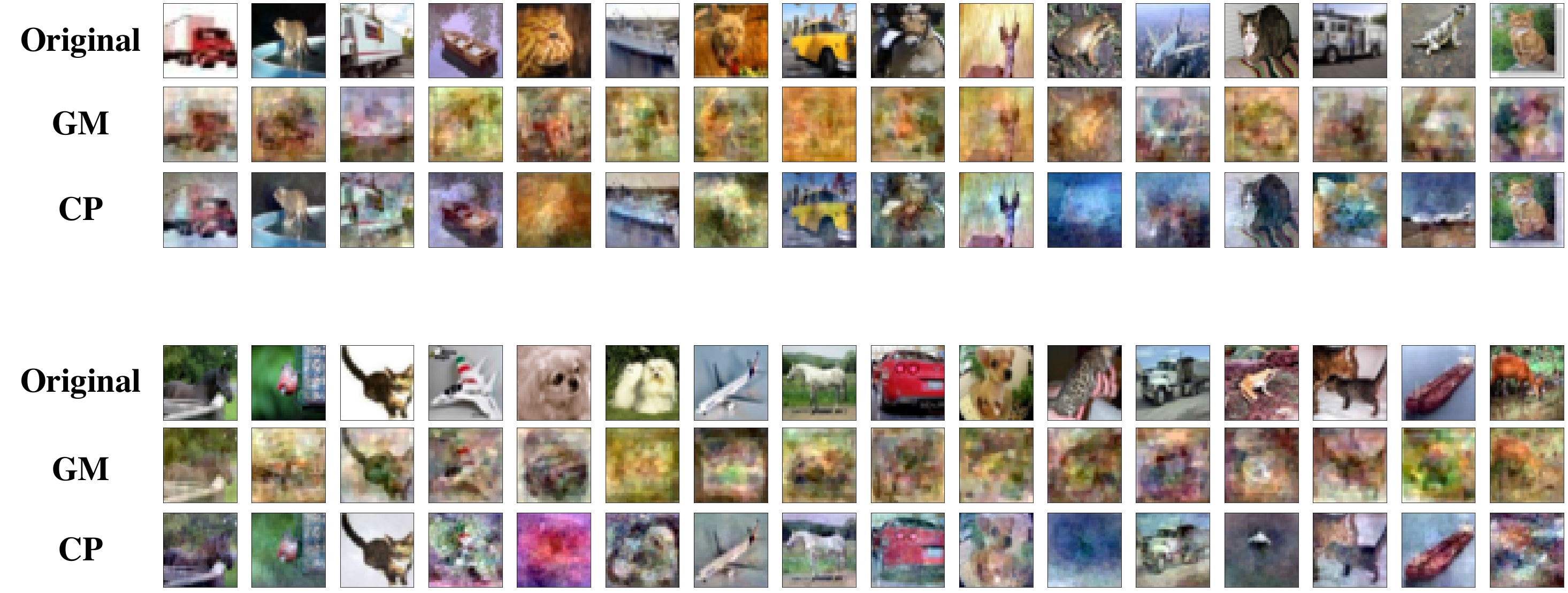, width=\textwidth}}
    \caption{Comparison of a random subset of images recovered by various gradient inversion attacks carried out using the gradients from FC2 with a batch of 64 CIFAR-10 images. Images are not cherry-picked.}
    \label{fig:cifar}
\end{figure*}}

\end{document}